\documentclass[11pt,letterpaper]{article}
\usepackage{ijcnlp2017}
\usepackage{times}
\usepackage{latexsym}
\usepackage{graphicx}
\usepackage{amsmath}
\usepackage{amsfonts}
\usepackage{comment}
\usepackage{multirow}
\usepackage{array}
\usepackage{subcaption}
\usepackage{hyperref}
\usepackage[font=small,labelfont=bf]{caption}

\ijcnlpfinalcopy

\def\ijcnlppaperid{***}

\title{What matters in a transferable neural network model for relation classification in the biomedical domain?}

\author{Sunil Kumar Sahu \and Ashish Anand \\
	Department of Computer Science and Engineering  \\
    Indian Institute of Technology Guwahati \\
    Assam, India \\
  {\tt \{sunil.sahu, anand.ashish\}@iitg.ernet.in}}

\date{}

\begin{document}

\maketitle

\begin{abstract}
Lack of sufficient labeled data often limits the applicability of advanced machine learning algorithms to real life problems. However efficient use of {\it Transfer Learning} (TL) has been shown to be very useful across domains. TL utilizes valuable knowledge learned in one task ({\it source task}), where sufficient data is available, to the task of interest ({\it target task}). In biomedical and clinical domain, it is quite common that lack of sufficient training data do not allow to fully exploit machine learning models. In this work, we present two unified recurrent neural models leading to three transfer learning frameworks for relation classification tasks. We systematically investigate effectiveness of the proposed frameworks in transferring the knowledge under multiple aspects related to source and target tasks, such as, similarity or relatedness between source and target tasks, and size of training data for source task. Our empirical results show that the proposed frameworks in general improve the model performance, however these improvements do depend on aspects related to source and target tasks. This dependence then finally determine the choice of a particular TL framework.

\end{abstract}

\section{Introduction}

Recurrent neural network (RNN) and its variants, such as long short term memory (LSTM) network have shown to be an effective model for many natural language processing tasks \cite{mikolov10,Graves13,karpathy2014,zhang2015,ChiuN15,zhou2016}. However, the requirement of huge gold standard labeled dataset for training makes it difficult to apply them on low resource tasks such as in biomedical domain. In the biomedical domain, obtaining labeled data not only is time consuming and costly but also requires domain knowledge. Transfer learning (TL) has been used successfully in such scenario across multiple domains. The aim of transfer learning is to store the knowledge gained while training a model for a Task-A ({\it Source Task}), where we have sufficient gold standard labeled data, and apply it to a different Task-B ({\it Target Task}) where we do not have enough training data \cite{pan2010}. 
In literature various TL frameworks have been proposed \cite{pan2010,mou2016,yosinski2014}. With the recent surge in applications of TL using neural network based models in computer vision and image processing~\cite{yosinski2014,azizpour2015} as well as in NLP~\cite{mou2016,zoph2016,zhilin2017}, this work explores TL frameworks using neural model for relation classification in biomedical domain. 

A very common approach to apply TL is to train learning model on source and target tasks in sequence. We refer to this approach as {\it sequential TL}. Further if there exists a bijection mapping between label sets of source and target tasks, then the entire model trained on source task can be transferred to the target task, otherwise only partial model can be utilized. In NLP, transferring of feature representation is the most common form of partial model transfer. Instead of performing the training in sequential manner, alternative way could be training the model on both source and target data simultaneously~\cite{zhilin2017}. This would be very similar to the {\it multi-task learning}~\cite{collobert08}. This way of simultaneous training can be done in multiple ways. These options give possibilities to design several variants of TL framework.

Apart from the options of using training data in different ways, using partial or complete model transfer, and presence or absence of bijection mapping between two label sets, other aspects such as {\it selection of source task}, {\it its size} and {\it relatedness or similarity with the target task} determine the selection of relevant TL model. Intuitively, it is preferred to have source task as much similar to the target task as we can obtain. For example, if target task is of binary classification of drug-drug interaction (DDI) mentioned in social media text or in doctor's notes, then we should look for the source task of binary classification of DDI mentioned in research articles. Here, the difference lie in the nature of texts appearing in the two corpora. In the first case of doctor's notes, text is likely to be short and precise compared to the research articles. In other words feature spaces representing data for source and target tasks differ from each other, although the two label sets are same. On the other hand, it is also possible that there does not exist any bijection between labels of source and target tasks. We can modify our previous example by making the target task as multi-class classification of DDI, to illustrate one such possible scenario.

Given that the various possibilities are arising in light of above discussion, we present two LSTM based models and corresponding three different TL frameworks in this study. Our motivation is to systematically explore various TL frameworks for the task of relation classification in biomedical domain and try to empirically analyze answers to few relevant questions. Our contribution can be summarized as follows:
 \begin{itemize}
  \item We present and evaluate three TL framework variants based on LSTM models for different relation classification tasks of biomedical and clinical text.
  \item We analyze effect of relatedness (implicit or explicit) between source and target tasks on the effectiveness of TL framework.
  \item We also explore how the size of the training data corresponding to source task affects effectiveness of TL frameworks.
 \end{itemize}

\section{Model Architectures}
\label{sect:models}
In this section we first explain a generic architecture of LSTM for relation classification task. Then we explain three ways of using this architecture for transferring knowledge from source task to target task. We assume that relation exists between two entities, referred to as {\it target entities}, positions of whom within the sentence are known.

The generic architecture of the neural network for relation classification task can be described in following layers: {\it word level feature layer}, {\it embedding layer}, {\it sentence level feature extraction layer}, {\it fully connected and softmax layers}. We define features for all words in {\it word level feature layer}, which also includes some features relative to the two targeted entities. In {\it embedding layer} every feature gets mapped to a vector representation through a corresponding embedding matrix. Raw features are combined from entire sentence and a fixed length feature representation is obtained in the {\it sentence level feature extraction layer}. Although a convolution neural network (CNN) or other variants of recurrent neural network can be used in this layer, we use bidirectional LSTM
because of its relatively better ability to take into account discontiguous features. {\it Fully connected and softmax layer} map thus obtained 
sentence level feature vectors to class probability. In summary, input for these models would be a sentence with the two targeted entities and output would be a probability distribution over each possible relation class between them.

 
 
\begin{figure*}[htb]
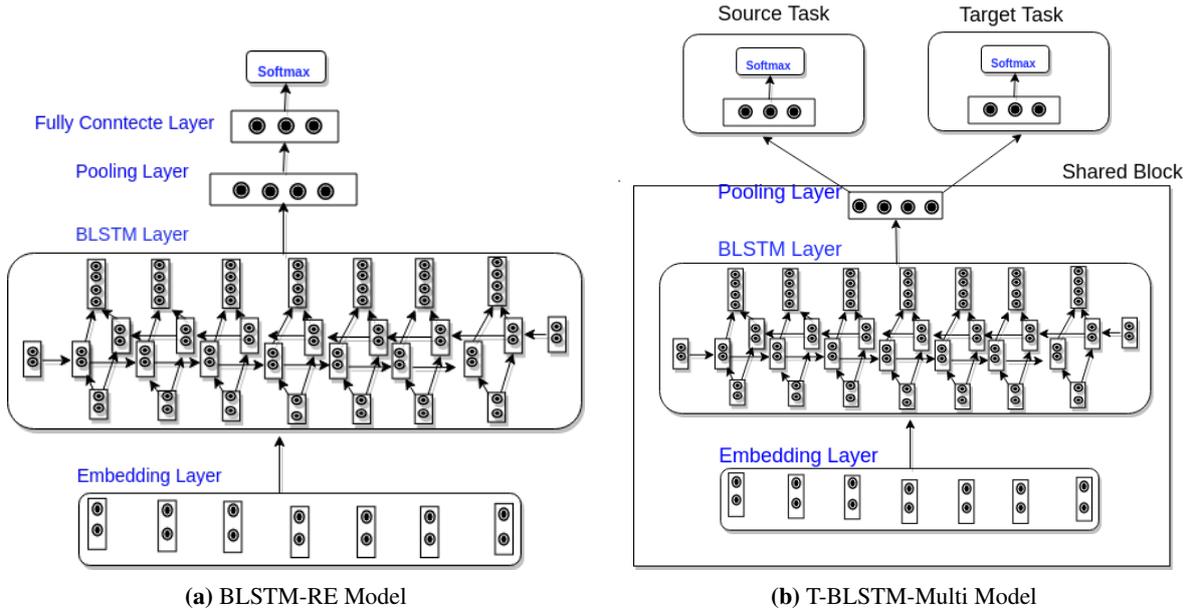

\centering
  \begin{subfigure}[b]{.5\linewidth}
    \centering
    \includegraphics[width=0.95\textwidth]{lstm_re.png}
    \caption{BLSTM-RE Model}
    \label{fig:model1a}
  \end{subfigure}%
  \begin{subfigure}[b]{.5\linewidth}
    \centering
    \includegraphics[width=0.95\textwidth]{tlstm_re.png}
    \caption{T-BLSTM-Multi Model}
    \label{fig:model1b}
  \end{subfigure}%
  \caption{Proposed Model Architect: BLSTM is bidirectional long short term memory network}
\end{figure*}
 
\subsection{BLSTM-RE}
Suppose $w_{1} w_{2} ..... w_{m}$ is a sentence of length $m$. Two targeted entities $e_{1}$ and $e_{2}$ corresponds to some words (or phrases) $w_i$ and $w_j$ respectively. In this work we use word and its position from both targeted entities as features in word level feature layer. Position features are important for relation extraction task because they let model to know the targeted entities~\cite{collobert11a}.  Output of {\it embedding layer} would be a sequence of vectors $x_{1} x_{2} ..... x_{m}$ where $ x_{i} \in \mathbb{R}^{(d_1+d_2+d_3)}$ is concatenation of word and position vectors. $d_1$,$d_2$ and $d_3$ are embedding lengths of word, position from first entity and position from second entity respectively. We use bidirectional LSTM with {\it max} pooling in the {\it sentence level feature extraction layer}. This layer is responsible to get optimal fixed length feature vector from entire sentence. Basic architecture is shown in the figure~\ref{fig:model1a}. We omit the mathematical equations as there is no modifications made in the standard bi-directional LSTM model \cite{Graves13}.

\subsection{{\it T-BLSTM-Mixed}}
{\it {\it T-BLSTM-Mixed}} is specific way to use {\it BLSTM-RE} model in transfer learning framework. In this case, instances from both source and target tasks are fed into the same {\it BLSTM-RE} model. While training we pick one batch of data from source or target in random order with equal probability. Since training is happening simultaneously for both source and target dataset, we can say that model will learn features which is applicable for both. It is quite obvious that this model is applicable for only those cases where bijection mapping between labels of source and target tasks exists. 

\subsection{{\it T-BLSTM-Seq}}
Convergence of neural network based models depends on initialization of model parameters. Several studies~\cite{hinton2006,bengio2007,collobert11a} have shown that initializing parameters with other supervised or unsupervised pre-trained model's value often improves the model convergence. In this framework of transfer we first train our model with source task's dataset and use the learned parameters to initialize model parameters for training target task in separate model. We call this framework as {\it {\it T-BLSTM-Seq}}. {\it {\it T-BLSTM-Seq}} can be applicable for both {\it same label set} as well as {\it disparate label set} transfer. We transfer entire network parameters if there exists bijection between source and target label sets, otherwise we only share model parameters up to the second last layer of the network. The left out last layer is randomly initialized.

\subsection{{\it T-BLSTM-Multi}}
We propose another transfer learning framework, called as {\it T-BLSTM-Multi}, using the same backbone of {\it BLSTM-RE} model. As shown in the figure~\ref{fig:model1b} this model has two {\it fully connected} and {\it softmax} layers, one for source task and other is for target task. Other layers of the models are shared for the two tasks. While training, parameters of the shared block get updated with training instances from both source and target data and {\it fully connected} layer gets updated only with its corresponding task data. Batch of instances are picked in similar manner as {\it {\it T-BLSTM-Mixed}}. This way of training is also called {\it multi-task learning} but in that case focus is on both source and target task performance. {\it {\it T-BLSTM-Multi}} model is also applicable for both {\it disparate label set} as well as {\it same label set} transfer.     

\subsection{Training and Implementation}
Pre-trained word vectors are used for initializing word embeddings and random vectors are used for other feature embedding. We use GloVe \cite{pennington14} on Pubmed corpus for obtaining word vectors. Dimensions of word and position embeddings are set to $100$ and $10$ respectively. Adam optimization \cite{adam2014} is used for training all models. All parameters i.e., word embedding, position embeddings and network parameters are being updated during training. We fixed batch size to $100$ for all the experiments. In case of {\it {\it T-BLSTM-Mixed}} and {\it {\it T-BLSTM-Multi}} every time we sample one task from source and target task based on binomial distribution, we set binomial probability to half and pick one batch from that task data for training. All the remaining hyperparameters are set according to \cite{sunil2017}. Entire implementation is done in Python language with {\it TensorFlow}\footnote{https://www.tensorflow.org/} library.

\section{Task Definitions and Used Datasets}
\label{sec:datasets}
In this section we briefly describe the tasks and corresponding datasets used in this study. Statistics of these datasets are given in Table \ref{tab:tab1}.

{\bf Drug Drug Interaction Extraction (DDI):} Drug-drug interaction is a state in which two or more drugs when given to a patient at the same time lead to undesired effects. Identifying DDI present in text is a kind of relation classification task, where given two drugs or pharmacological substances in a sentence we need to classify whether there is an interaction between the two or not. In {\it Ex.1} drug {\it Lithium} and {\it Diuretics} are interacting because they are advised to be not given together.

{\bf Ex.1}: {\it Lithium$_{drug}$ generally should not be given with diuretics$_{drug}$}

{\bf Drug Drug Interaction Class Extraction (DDIC):} DDI can appear in text with different semantic senses, which we call as  DDI class. In case of DDIC, we need to identity exact class of interactions among drugs in the sentence. For instance in example {\it Ex.1} type of interaction is {\it advise} as advice is being given.  In SemEval 2013\footnote{https://www.cs.york.ac.uk/semeval-2013/task9/} DDI Extraction task had $4$ kinds of interaction {\it Advise, Effect, Mechanism} and {\it Int}. 

{\bf Adverse Drug Event Extraction (ADE):} Adverse drug event is the condition in which an adverse effect happens due to consumption of a drug. In NLP, ADE extraction is the process of extracting adverse relation between a drug and a condition or disease in text. For instance in {\it Ex.2} for treating patient suffering from {\it thyrotoxicosis} disease with {\it methimazole} has led to adverse effect. 

{\bf Ex.2}: {\it A 43 year old woman who was treated for thyrotoxicosis$_{Disease}$ with methimazole$_{Drug}$ developed agranulocytosis}

{\bf Event Argument Extraction (EAE):} In biomedical domain event is broadly described as a change on the state of a bio-molecule or bio-molecules \cite{pyysalo2012}. Every events have their own set of arguments and EAE is the task of identifying all arguments of an event and their roles. In this task we have entities and triggers (representing an event) present in a sentence are given and the task is to find the role (relation) between all pairs of triggers and entities. For this work we don't differentiate between different types of role. This implies that if an entity is a argument of a trigger then there is positive relation between them otherwise negative. For instance in {\it Ex.3} {\it (reptin, regulates)}, {\it (regulates, growth)} and {\it (growth, heart)} have positive relation.

{\bf Ex.3}: {\it Reptin$_{Protein}$ regulates$_{Regulation}$ the growth$_{Growth}$ of the heart $_{Organ}$}.

{\bf Clinical Relation Extraction (CRE):} Clinical relation extraction is the task of identifying relation among clinical entities such as {\it Problem, Treatment and Test} in clinical notes or discharge summaries. In {\it Ex.4} {\it allergic} and {\it rash} have {\it problem improve problem} relation. 

{\bf Ex.4}: {\it She is allergic$_{Problem}$ to augmentin which gives her a rash$_{Problem}$.}

\begin{table}[ht]
\centering
\scalebox{0.7}{
\begin{tabular}{|c|c|c|c|}
\hline
{\textbf{Task}} & {\textbf{Corpus}} & {\textbf{Training Set}} & {\textbf{Test Set}}  \\ \hline
\multirow{3}{*}{BankDDI} & 
Pairs  			& 14176  &  3694 \\ \cline{2-4}
& Positive DDIs	& 3617   &  884 \\ \cline{2-4}	
& Negative DDIs	& 11559  & 2810    \\  
\hline
\hline
\multirow{3}{*}{MedDDI} & 
Pairs  			& 1319  &  334\\ \cline{2-4}
& Positive DDIs	& 227   &  95 \\ \cline{2-4}	
& Negative DDIs	& 1092  &  239\\  
\hline
\hline
\multirow{7}{*}{BankDDIC} & 
Pairs  			& 14176 & 3694\\ \cline{2-4}
& Negative DDIs	& 11559 & 2810\\ \cline{2-4}	
& Effect 		& 1471  & 298 \\ \cline{2-4}	
& Mechanism 	& 1203  & 278 \\ \cline{2-4}	
& Advise 		& 813   & 214 \\ \cline{2-4}	
& Int 			& 130   & 94  \\ 
\hline
\hline
\multirow{7}{*}{MedDDIC} & 
Pairs  			& 1319  &  334\\ \cline{2-4}
& Negative DDIs	& 1092  &  239\\ \cline{2-4}	
& Effect 		& 149   &  62 \\ \cline{2-4}	
& Mechanism 	& 61  	&  24 \\ \cline{2-4}	
& Advise 		& 7   	&  7  \\ \cline{2-4}	
& Int 			& 10   	&  2  \\ \cline{2-4}	
\hline
\hline
\multirow{3}{*}{ADE} &
Pairs  			& 8867 & 3802 \\ \cline{2-4}
& Positive ADEs	& 4177 & 1791 \\ \cline{2-4}	
& Negative ADEs	& 4690 & 2011 \\ 
\hline
\hline
\multirow{3}{*}{EAE} &
Pairs  			& 21594  &  11443 \\ \cline{2-4}	
& Positive EAEs	& 4492   &  2202  \\ \cline{2-4}	
& Negative EAEs	& 17102  &  9241  \\ 
\hline
\hline
\multirow{7}{*}{CRE} &
Pairs  			&  43602 & 18690    \\ \cline{2-4}		
& Negative CREs	&  36324 & 15995  	\\ \cline{2-4}
& TeRP 	& 2136  & 915	\\ \cline{2-4}
& TrAP 	& 1832 	& 784	\\ \cline{2-4}
& PIP 	& 1541 	& 660 	\\ \cline{2-4}
& TrCP 	& 368  	& 157	\\ \cline{2-4}
& TeCP 	& 353	& 150	\\ \hline
\end{tabular}
}
\caption{Statistics of all dataset}
\label{tab:tab1}
\end{table}

\subsection{Source Data}
{\bf BankDDI}: It is manually annotated dataset for DDI extraction, collected from Drug Bank\footnote{https://www.drugbank.ca/} documents. Drug Bank contains drug information in the form of documents which has been written by medical practitioners. We collected this dataset from SemEval 2011 DDI extraction challenge \cite{segura2011}. 

{\bf BankDDIC}: This dataset is same as BankDDI with task of DDI class recognition \cite{segura2013}.

{\bf ADE}: ADE extraction dataset was collected from  \cite{gurulingappa2012data,gurulingappa2012}. The shared dataset contains manually annotated 
adverse drug events mentioned in a corpus of Medline abstracts.

{\bf EAE}: We used MLEE\footnote{http://nactem.ac.uk/MLEE/} corpus for event argument identification \cite{pyysalo2012}. MLEE dataset has $20$ types of events trigger and $11$ entity types for relation classification. 

{\bf CRE}: For clinical relation classification, we collected dataset from i2b2 2010\footnote{https://www.i2b2.org/NLP/Relations/} clinical information extraction challenge \cite{uzuner10a}. We consider $TrCP$, $TrAP$, $PIP$, $TeRP$ and $TeCP$ classes in this case. 

\subsection{Target Data}
{\bf MedDDI}: MedDDI is manually annotated dataset for DDI extraction, collected from MedLine abstract. This dataset was also shared as part of SemEval-2013 DDI extraction challenge. There are several ways MedDDI is different from BankDDI: MedDDI was collected from MedLine abstracts whereas BankDDI is from DrugBank documents. MedLine abstracts, as being a part of research articles, contains lots of technical terms and usually sentences are longer. On the other hand DrugBank documents contains concise, relatively smaller and easily comprehensible sentences written by medical practitioners. In BankDDI and MedDDI datasets, we removed few negative instances based on same rules used in \cite{sunil2017,zhao2016}.

{\bf MedDDIC}: It is same dataset with labels includes exact class of interaction. Both this dataset was used in SemEval-2013 DDI extraction task.

{\bf CRE$_5$}: In this case we take $5\%$ from each class of CRE's training dataset and considered that as training set. The test set remains same.

\subsection{Preprocessing }
We use same preprocessing strategies for all datasets. Pre-processing steps include, all words were converted into lower case form, sentences were tokenized with geniatagger\footnote{http://www.nactem.ac.uk/GENIA/tagger/}, digits were replaced with $DG$ symbol. Further, if any sentence have more than two entities, we create a separate instance for every pair of entities and in all sentences two targeted entities were replaced with their types and position of entity. For example, sentence in {\it Ex.4} will become {\it She is ProblemA to augmentin which gives her a ProblemB}. The complete CRE and ADE datasets were separated into training and test sets by randomly selecting $30\%$ instances from each class as test set and remaining as training set.

\section{Results and Discussion}
\label{sec:results}
First we discuss our experiment design to evaluate performance of the three TL frameworks under various settings. Later we analyze and discuss obtained results.

We treat the performance of bidirectional LSTM model on target task as baseline. In baseline experiments, training was done on the training set of each of the three target data and performance on the respective test sets are then reported in Table~\ref{tab:baseline}.
\begin{table}[h]
\centering
\scalebox{0.9} {
\begin{tabular} {|l|c|c|c|}
\hline
\textbf{Task} & {\bf Precision} & {\bf Recall} & {\bf F Score}  \\ \hline
MedDDI			&  0.561$_{(0.03)}$ & 0.431$_{(0.03)}$& 0.488$_{(0.02)}$ \\ \hline
MedDDIC			& 0.684$_{(0.08)}$  & 0.273$_{(0.01)}$ & 0.390$_{(0.03)}$ \\ \hline
CRE$_5$			& 0.529$_{(0.04)}$  & 0.492$_{(0.01)}$ & 0.510$_{(0.009)}$  \\ \hline
\end{tabular}
}
\caption{Baseline Performance: Results of {\bf BLSTM-RE} model on the three different target tasks. Numbers in Precision, Recall and F Score column indicate result corresponding to best F1 Score and subscripts are standard deviation of five runs of model}
\label{tab:baseline}
\end{table}

\begin{table*}[t]
\begin{minipage}{\textwidth}
\centering
\scalebox{0.9}{
\begin{tabular}  
{|p{0.1\linewidth}|p{0.41\linewidth}|p{0.1\linewidth}p{0.1\linewidth}p{0.12\linewidth}p{0.09\linewidth}|} 
\hline
{\bf Type } & \textbf{Model} & {\bf Precision} & {\bf Recall} & {\bf F Score} & $\Delta$ \\ \hline
\multirow{6}{*}{\bf Similar}
& {\it T-BLSTM-Mixed}$_{(BankDDI\Rightarrow MedDDI)}$  & 0.656$_{(0.02)}$ & 0.705$_{(0.03)}$ & 0.680$_{(0.03)}$ & 39.34\% \\ 
& {\it T-BLSTM-Seq}$_{(BankDDI\Rightarrow MedDDI)}$  & 0.678$_{(0.02)}$ & 0.621$_{(0.03)}$ & 0.648$_{(0.03)}$ & 32.78\%  \\ 
& {\it T-BLSTM-Multi}$_{(BankDDI\Rightarrow MedDDI)}$ & $0.701_{(0.05)}$ & 	$0.568_{(0.05)}$ &	0.627$_{(0.02)}$ &28.48\% \\ \cline{2-6}
& {\it T-BLSTM-Mixed}$_{(BankDDIC\Rightarrow MedDDIC)}$& 0.631$_{(0.04)}$& 0.505$_{(0.02)}$ & 0.561$_{(0.01)}$ & 43.84\%  \\
& {\it T-BLSTM-Seq}$_{(BankDDIC\Rightarrow MedDDIC)}$& 0.600$_{(0.03)}$& 0.463$_{(0.02)}$ & 0.550$_{(0.01)}$ &  41.02\% \\
& {\it T-BLSTM-Multi}$_{(BankDDIC\Rightarrow MedDDIC)}$ & $0.579_{(0.01)}$ & $0.421_{(0.006)}$ & $0.487_{(0.006)}$ & 24.87\%\\ 
\hline
\multirow{6}{*}{\bf Dissimilar}
& {\it T-BLSTM-Mixed}$_{(ADE\Rightarrow MedDDI)}$ & 0.494$_{(0.02)}$ & 0.515$_{(0.03)}$ & 0.505$_{(0.02)}$ &  3.48\% \\ 
& {\it T-BLSTM-Seq}$_{(ADE\Rightarrow MedDDI)}$ & 0.595$_{(0.02)}$ & 0.294$_{(0.03)}$ & 0.394$_{(0.02)}$ & -19.26\%  \\ 
& {\it T-BLSTM-Multi}$_{(ADE\Rightarrow MedDDI)}$& $0.533_{(0.02)}$ &$0.505_{(0.02)}$ & $0.518_{(0.01)}$  & 6.14\% \\ \cline{2-6}
& {\it T-BLSTM-Mixed}$_{(EAE\Rightarrow MedDDI)}$ &0.540$_{(0.03)}$ & 0.557$_{(0.05)}$ & 0.549$_{(0.02)}$ & 12.5\%  \\  
& {\it T-BLSTM-Seq}$_{(EAE\Rightarrow MedDDI)}$ &0.544$_{(0.03)}$ & 0.515$_{(0.04)}$ & 0.529$_{(0.02)}$ &  8.40\% \\  
& {\it T-BLSTM-Multi}$_{(EAE\Rightarrow MedDDI))}$& $0.538_{(0.02)}$ & $0.589_{(0.05)}$ & $0.562_{(0.02)}$ & 15.16\%\\
\hline 
\end{tabular}
}
\caption{Results of TL frameworks in case of \textbf{same label set transfer}. Here ($X\Rightarrow Y$) indicates transferring from $X$ dataset to $Y$ dataset and $Type$ indicates nature of source and target dataset. Numbers in Precision, Recall and F Score column indicate result corresponding to best F1 Score and subscripts  are standard deviation of five runs of model. $\Delta$ is relative percentage improvement over baseline (without TL) method}
\label{tab:tl_res_same_label}
\end{minipage}
\end{table*}
Needless to mention that baseline model do use pre-trained word embeddings, a form of unsupervised transfer learning framework. As multiple studies have already 
shown the superior performance of such models using pre-trained vectors than using random vectors, we do not perform any experiment related to that. Five runs with different random initialization were taken for each model and the best result in terms of F1-score along with corresponding precision and recall are shown in Tables. We experiment with different combinations of source and target tasks to analyze the effect of similarity of feature spaces corresponding to source and target tasks as well as their label sets on the choice of TL frameworks.

\subsection{Performance on Same Label Set Transfer}
Let's first look at the relative improvement of various TL models over the baseline results on DDI and DDIC tasks. Table~\ref{tab:tl_res_same_label} shows the performance of all TL models on these two tasks under various settings. {\it Type} in the Table~\ref{tab:tl_res_same_label} indicates semantic relatedness of source and target tasks. For example, data in both {\it BankDDI} and {\it MedDDI} indicate drug-drug interaction, and hence are of the same semantic type. But {\it EAE} gives existence of trigger argument relation which is not of the same semantic type as drug-drug interaction, although both tasks fall into binary classification. As the results indicate, {\it T-BLSTM-Mixed} model gave the best performance (in terms of F1-score) for the {\it similar} type tasks, whereas {\it T-BLSTM-Multi} gave the worst. However all the TL models gave significant improvement over the baseline results. {\it T-BLSTM-Mixed} obtained approximately $40\%$ relative improvement over the baseline for the DDI task and approximately $44\%$ for the DDIC task. On the other hand, {\it T-BLSTM-Multi} gave best performance for the {\it dissimilar} type tasks and {\it T-BLSTM-Seq} gave the worst. In fact, {\it T-BLSTM-Seq} gave relatively poor performance than baseline in one case. 

\subsection{Performance on Disparate Label Set Transfer}
\begin{table*}[t]
\begin{minipage}{\textwidth}
\centering
\scalebox{0.9}{
\begin{tabular} 
{|p{0.23\linewidth}|p{0.1\linewidth}p{0.1\linewidth}p{0.1\linewidth}|p{0.1\linewidth}p{0.1\linewidth} p{0.1\linewidth}|} \hline
\multirow{2}{*}{\textbf{Source$\Rightarrow$Target}} & \multicolumn{3}{c|}{\textbf{{\it T-BLSTM-Seq}}} & \multicolumn{3}{c|}{\textbf{{\it T-BLSTM-Multi}}} \\ \cline{2-7}
 & Precision & Recall & F Score & Precision & Recall & F Score\\ \hline
BankDDI$\Rightarrow$MedDDIC & 0.50$_{(0.08)}$ & 0.378$_{(0.03)}$ & 0.431$_{(0.008)}$&  $0.603_{(0.05)}$ & $0.368_{(0.03)}$ 	& $0.457_{(0.03)}$ \\  
CRE$\Rightarrow$MedDDIC		& 0.448$_{(0.05)}$& 0.368$_{(0.03)}$& 0.404$_{(0.02)}$ &  $0.468_{(0.02)}$ & $0.389_{(0.02)}$ 	& $0.425_{(0.01)}$ \\  
EAE$\Rightarrow$MedDDIC		& 0.596$_{(0.04)}$& 0.326$_{(0.03)}$& 0.421$_{(0.02)}$&  $0.488_{(0.03)}$ & $0.452_{(0.06)}$	& $0.469_{(0.04)}$ \\ 
ADE$\Rightarrow$MedDDIC		& 0.512$_{(0.06)}$ & 0.221$_{(0.01)}$ & 0.308$_{(0.01)}$ &  $0.447_{(0.04)}$ & $0.400_{(0.03)}$ & $0.422_{(0.02)}$    \\ 
\hline
\hline
BankDDI$\Rightarrow$CRE$_5$ &0.555$_{(0.02)}$ & 0.485$_{(0.01)}$& 0.518$_{(0.01)}$& 0.546$_{(0.01)}$ & 0.508$_{(0.02)}$ & 0.526$_{(0.006)}$ \\ 
BankDDIC$\Rightarrow$CRE$_5$&0.564$_{(0.04)}$ & 0.447$_{(0.02)}$& 0.498$_{(0.01)}$ & 0.523$_{(0.01)}$ & 0.557$_{(0.02)}$ & 0.539$_{(0.007)}$\\ 
EAE$\Rightarrow$CRE$_5$& 0.543$_{(0.02)}$& 0.533$_{(0.02)}$ & 0.538$_{(0.006)}$ &  0.587$_{(0.01)}$ & 0.548$_{(0.01)}$ & 0.567$_{(0.01)}$ \\ 
ADE$\Rightarrow$CRE$_5$&0.516$_{(0.003)}$ & 0.503$_{(0.01)}$& 0.509$_{(0.006)}$ &  0.598$_{(0.03)}$ & 0.483$_{(0.02)}$ & 0.535$_{(0.007)}$   \\ 
\hline
\hline
BankDDIC$\Rightarrow$MedDDI & 0.605$_{(0.04)}$ & 0.452$_{(0.03)}$ & 0.518$_{(0.01)}$& $0.623_{(0.04)}$ & $0.557_{(0.03)} $ & $0.588_{(0.02)}$ \\
CRE$\Rightarrow$MedDDI & 0.569$_{(0.11)}$ & 0.473$_{(0.17)}$ & 0.517$_{(0.03)}$ & $0.631_{(0.05)}$ & $0.505_{(0.02)}$ & $0.561_{(0.02)}$ \\ 
\hline
\end{tabular}
}
\caption{Results of {\it T-BLSTM-Seq} and {\it T-BLSTM-Multi} on \textbf{disparate label set transfer} task. Here ($X\Rightarrow Y$) indicates transferring from $X$ dataset to $Y$ dataset. Numbers in Precision, Recall and F Score column indicate result corresponding to best F1 Score and subscripts  are standard deviation of $5$ runs of model.}
\label{tab:tl_res_disparate_label}
\end{minipage}
\end{table*}
Next we examine performance of relevant TL models when there does not exist a bijection between the two label sets corresponding to source and target tasks (Table \ref{tab:tl_res_disparate_label}). As {\it T-BLSTM-Mixed}, by design, require the existence of bijection between the two label sets, we exclude this model for this case. Among the rest two, {\it T-BLSTM-Multi} always led to significantly improved performance compared to the respective baseline results. On the other hand, performance of the {\it T-BLSTM-Seq} model is not so consistent specially when {\it ADE} was used as source data. 

\subsection{Analyzing Similarity between Source and Target Tasks}
\label{sec:similarity}
Earlier we observe that similarity between source and target tasks affects the relative performance of each TL framework. Let us try to analyze this observation. {\it T-BLSTM-Mixed} and {\it T-BLSTM-Seq} models transfer full knowledge or in other words, both model share the complete model between source and target tasks. This allow the last layers to see more examples and to be adaptive to samples from both source and target data. On the other hand, {\it T-BLSTM-Multi} only share the partial model upto the second last layers. In this case, last layer for source and target tasks are trained separately and is not being shared between the two. Thus the last layers are being specific to the respective tasks.  When there is similarity between source and target tasks, as well as there exists a bijection, target tasks gets benefited by sharing full model. In such scenario co-training seems better suited as {\it T-BLSTM-Mixed} was found to be the best among three frameworks. On the other hand, {\it T-BLSTM-Multi} fail to exploit the full knowledge present in the training data of source task. But this becomes advantageous for the {\it T-BLSTM-Multi} framework in case of absence of bijection between source and target label sets. The last layer, in {\it T-BLSTM-Multi}, takes the shared knowledge and tune it into the specific target task. This observation also fits well with the observations made earlier in \cite{yosinski2014} that the initial layers are relatively generic and become more specific as we go towards the last layer.

\begin{figure}[h]
\centering
  \begin{subfigure}[b]{0.8\linewidth}
    \includegraphics[width=1\textwidth]{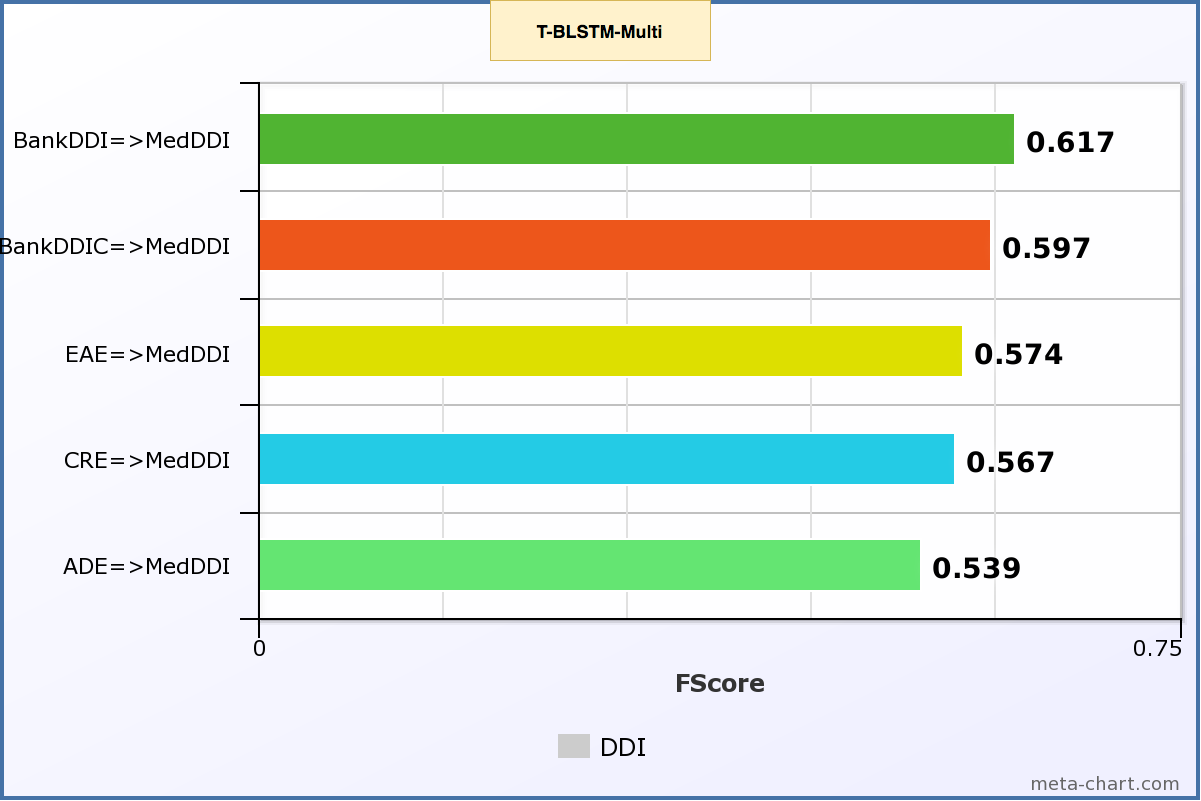}
    \caption{T-BLSTM-Multi Model (DDI Extraction)}
    \label{fig:1a}
  \end{subfigure}%
  
  \begin{subfigure}[b]{0.8\linewidth}
    \includegraphics[width=1\textwidth]{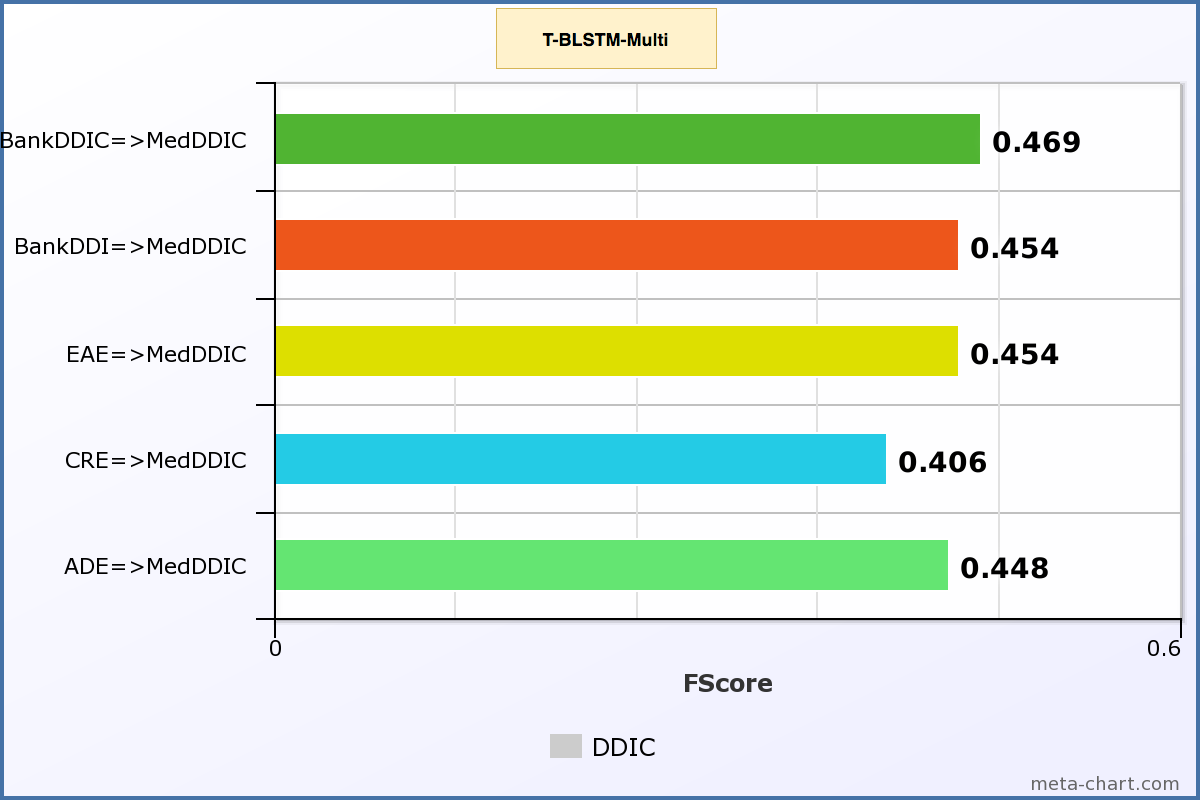}
    \caption{T-BLSTM-Multi : Model (DDIC Extraction)}
    \label{fig:1b}
  \end{subfigure}%
    \caption{Performance of proposed models with different source task on same size data}
    \label{fig:res_same_source_size}
\end{figure}
Variability in training set size of different source data could have influenced the observed performance difference. Hence to take this effect out from the consideration, all source data was made of the same size ($8867$) as the minimum among all source training data. During random selection, proportion of instances from all classes were maintained as in the original set. We have shown only the results obtained by {\it T-BLSTM-Multi} in Figure~\ref{fig:res_same_source_size} but similar results are obtained for other models as well. We observe the performance obtained from using different source data but of same size match with performance obtained with the same set of source data but of different sizes. This indicates that context and label mapping played more crucial role than size of selected source data.


\subsection{Analyzing Size of Source Task Dataset } 
\label{size_analysis}
\begin{figure}[h]
\centering
  \begin{subfigure}[b]{.8\linewidth}
    \includegraphics[width=1\textwidth]{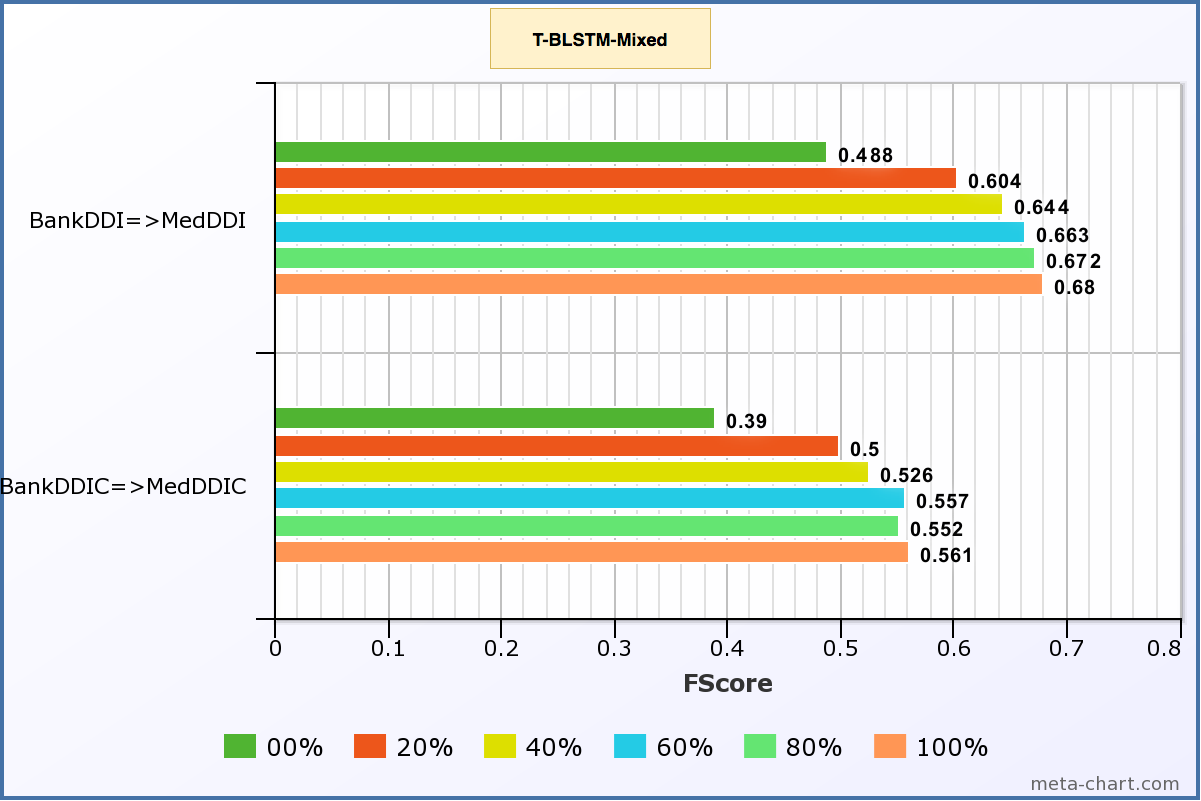}
    \caption{T-BLSTM-Mixed : transfer on similar task}
  \label{fig:source_size_sim}
  \end{subfigure}%
  
  \begin{subfigure}[b]{.8\linewidth}
    \includegraphics[width=1\textwidth]{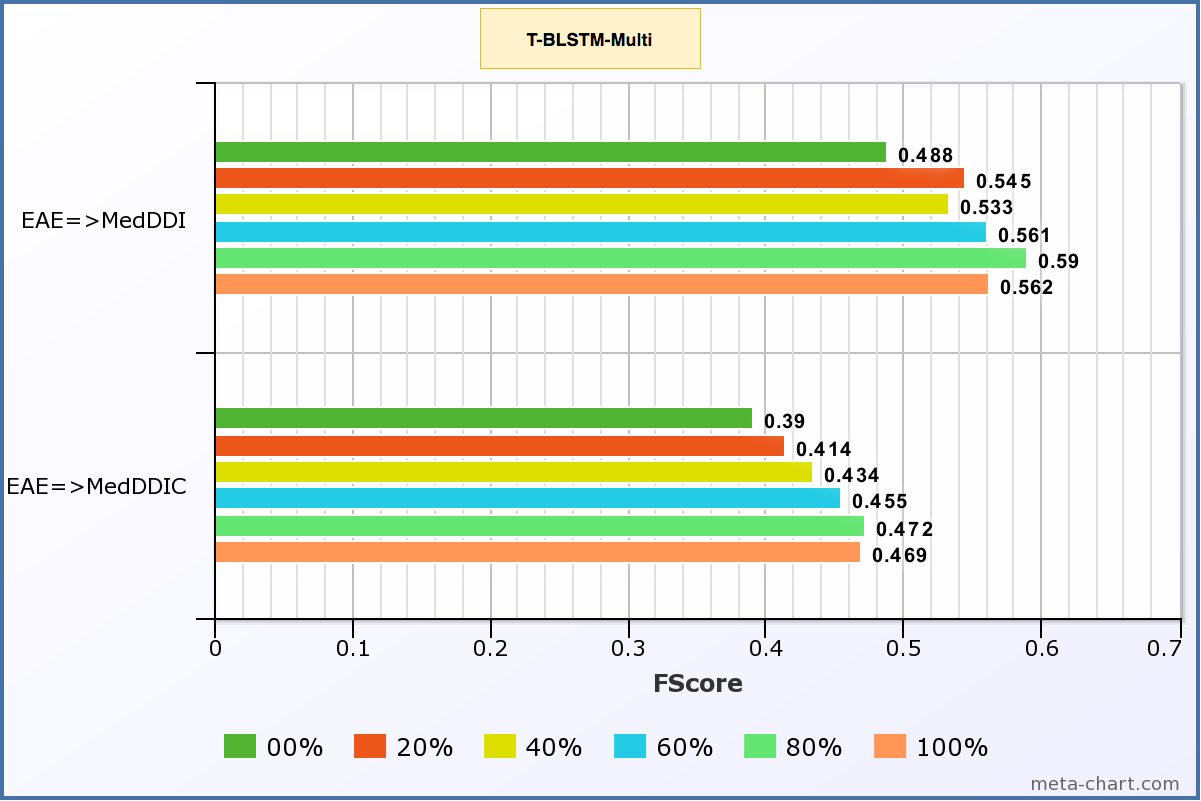}
    \caption{T-BLSTM-Multi : transfer on dissimilar task}
    \label{fig:source_size_dis}
  \end{subfigure}%
  \caption{Performance of proposed models with different training set size corresponding to source task}
  \label{fig:source_size}
\end{figure}

One of the important arguments generally given for the use of transfer learning is insufficient dataset for the target task hinders the performance of learning algorithms. Performance can be enhanced by utilizing information available in relatively higher amount of source data. In this section we investigate the effect of different size of source data on the performance improvement of the {\it T-BLSTM-Mixed} and {\it T-BLSTM-Multi} models. Figure~\ref{fig:source_size} shows the results on both similar and dissimilar tasks. In both scenario, even having $20\%$ of source data significantly improves the performance. However, there is a consistent increasing trend in improvement observed for the similar tasks, whereas performance was found to be little fluctuating for the dissimilar tasks. The fluctuation could be due to too much of source data may confuse the model.

\subsection{Comparison with State-of-art Results}
\label{compare}
At the end we compare our results with the state-of-the-art results obtained on target tasks of {\it DDI} and {\it DDIC}. Table \ref{tab:ddi_res_comp} shows best results of the SemEval 2010 DDI extraction challenge \cite{segura2013} as well as the results obtained by BLSTM-RE and {\it T-BLSTM-Mixed} models on dissimilar tasks. We can observe that although BLSTM-RE can not outperform the best results of the challenge but under the TL framework, {\it T-BLSTM-Multi} even using dissimilar tasks improved the state-of-the-art results.
\begin{table}[h]
\centering
\scalebox{0.8} {
\begin{tabular} {|l|c|c|}
\hline
\textbf{Models} & {\bf DDIC} & {\bf DDI}  \\ \hline
FBK-Irst\cite{chowdhury2013} 	& 0.398 & 0.530 \\ \hline
SCAI\cite{bobicscai} 			& 0.420	& 0.47	\\ \hline
WBI\cite{thomas2013}			& 0.365	& 0.503 \\ \hline
UTurku\cite{bjorne2013}			& 0.286	& 0.479 \\ \hline
UMAD\cite{rastegar2013}			& 0.312	& 0.479 \\ \hline
\hline
BLSTM-RE			 			& 0.390	& 0.488 \\ \hline
{\it T-BLSTM-Multi}$_{EAE=>\bigoplus}$	& {\bf 0.469}	& {\bf 0.562} \\ \hline
{\it T-BLSTM-Multi}$_{CRE=>\bigoplus}$	& 0.425 & 0.561 \\ \hline
{\it T-BLSTM-Multi}$_{ADE=>\bigoplus}$	& 0.422 & 0.518\\ \hline 
\end{tabular}
}
\caption{Performance comparison of existing methods for DDI and DDIC task. Here values indicate {\bf F1 Score} of both task. $\bigoplus$ is MedDDI or MedDDIC}
\label{tab:ddi_res_comp}
\end{table}

\section{Related Work}
\label{sec:rel_works}
Recurrent neural network and its variants have been successfully applied to many semantic relation classification tasks. Authors in \cite{zhang2015,shu2015} have used recurrent neural network with max pooling operation while \citet{zhou2016} used RNN with attentive pooling for the same relation classification task. \citet{LiZFQJ16} partitioned the sentence with targeted entities and a separate RNN was trained to obtain division specific features and use them for classification. 
 \citet{LiZFQJ16} used recurrent neural network for semantic relations as well as Bio Event argument relation extraction tasks. Although none of these works have
 used transfer learning but have used RNN models for various relation classification tasks.

The proposed TL frameworks are closely related to the works of \cite{zhilin2017,mou2016,collobert08}. \cite{zhilin2017} have introduced variety of TL
frameworks using gated recurrent neural network (GRU). They have evaluated the proposed frameworks on different sequence labeling tasks, such as {\it PoS tagging} and {\it chunking}. \citet{mou2016}, similar to the study by \citet{yosinski2014} for image processing tasks, evaluated CNN and RNN based TL frameworks for sentence classification and sentence pair modeling tasks. \citet{collobert08} have used window based neural network and convolution neural networks for several sequence labeling tasks in the multi-task learning framework. \citet{zoph2016} have explored transfer learning for neural machine translation tasks. They have shown significant improvement in many low resource language translation tasks. Their model repurpose the learned model, trained on high resource language translation dataset (source task), for target task.

\section{Conclusions} 
 \label{sec:conc}
In this work we present various transfer learning frameworks based on LSTM models for relation classification task in biomedical domain. We observe that in general transfer learning do help in improving the performance. However, similarity of source tasks with the target task as well as size of corresponding source data affects the performance and hence plays important role in selection of appropriate TL framework.

\bibliography{emnlp2017}
\bibliographystyle{ijcnlp2017}

\end{document}